\documentclass{article}

\usepackage{arxiv}
\usepackage[utf8]{inputenc} % allow utf-8 input
\usepackage[T1]{fontenc}    % use 8-bit T1 fonts
\usepackage{hyperref}       % hyperlinks
\usepackage{url}            % simple URL typesetting
\usepackage{booktabs}       % professional-quality tables
\usepackage{amsfonts}       % blackboard math symbols
\usepackage{nicefrac}       % compact symbols for 1/2, etc.
\usepackage{microtype}      % microtypography
\usepackage{lipsum}
\usepackage{lscape} 

\usepackage{epsfig} % for postscript graphics files
\usepackage{amsmath} % assumes amsmath package installed
\usepackage{amssymb}  % assumes amsmath package installed
\usepackage{multirow}

\title{Gated Variational AutoEncoders: \\ Incorporating Weak Supervision to Encourage Disentanglement}

\usepackage[nolist]{acronym}
\usepackage{float}
\acrodefplural{vae}[VAEs]{Variational AutoEncoders}
\acrodefplural{gan}[GANs]{Generative Adversarial Networks}

\begin{acronym} 
\acrodef{vae}[VAE]{Variational AutoEncoder}
\acrodef{gan}[GAN]{Generative Adversarial Network}
\end{acronym}

\author{
  Matthew J. Vowels\\
  m.j.vowels@surrey.ac.uk \\
   \And
 Necati Cihan Camgoz \\
 n.camgoz@surrey.ac.uk \\
  \And
  Richard Bowden\\
  r.bowden@surrey.ac.uk
  \And
  \\
  Centre for Vision, Speech, and Signal Processing \\ University of Surrey, United Kingdom
}

\begin{document}
\maketitle

\begin{abstract}
\acp{vae} provide a means to generate representational latent embeddings. Previous research has highlighted the benefits of achieving representations that are disentangled, particularly for downstream tasks. However, there is some debate about how to encourage disentanglement with \acp{vae} and evidence indicates that existing implementations of \acp{vae} do not achieve disentanglement consistently. The evaluation of how well a VAE's latent space has been disentangled is often evaluated against our subjective expectations of which attributes should be disentangled for a given problem. Therefore, by definition, we already have domain knowledge of what should be achieved and yet we use unsupervised approaches to achieve it. We propose a weakly-supervised approach that incorporates any available domain knowledge into the training process to form a Gated-VAE. The process involves partitioning the representational embedding and gating backpropagation. All partitions are utilised on the forward pass but gradients are backpropagated through different partitions according to selected image/target pairings. The approach can be used to modify existing VAE models such as $\beta$-VAE, InfoVAE and DIP-VAE-II. Experiments demonstrate that using gated backpropagation, latent factors are represented in their intended partition. The approach is applied to images of faces for the purpose of disentangling head-pose from facial expression. Quantitative metrics show that using Gated-VAE improves average disentanglement, completeness and informativeness, as compared with un-gated implementations. Qualitative assessment of latent traversals demonstrate its disentanglement of head-pose from expression, even when only weak/noisy supervision is available.
\end{abstract}

% keywords can be removed
\keywords{Variational Autoencoders \and Disentanglement \and Representation Learning}

\section{INTRODUCTION}
Variational AutoEncoders (\acp{vae}) have gained in popularity for the unsupervised generation of low-dimensional representational embeddings over high-dimensional distributions such as images \cite{infovae, tshannen, sonderby}. It has been demonstrated \cite{higgins, infovae,kumar2, bengio1} that if representations are disentangled (such that each representational dimension uniquely and independently corresponds with a single generative factor), then better results are achieved in downstream tasks. However, how to achieve disentanglement is an ongoing area of research, and there is evidence that recent proposals do not achieve disentanglement consistently \cite{dai3, locatello}.

We present a novel weakly-supervised approach to training \acp{vae} which we will refer to as a Gated-VAE (see Figure \ref{fig:simplediagram}). This involves gating the backpropagation of gradients through a partitioned latent space, where the gating is determined by the input and target image pairs. The modification can be applied to any existing VAE model. The efficacy of the Gated-VAE is demonstrated quantitatively using the disentanglement, completeness and informativeness metrics from \cite{eastwood} and is compared against un-gated implementations of a \ac{vae}. We consider the relative importance of disentanglement versus informativeness by comparing the quantitative metrics thereof. We also qualitatively evaluate latent traversals for images of faces where noisy/weak supervision is available.

The paper is structured as follows: In Section \ref{sec:background} we provide an overview of recent endeavours to improve disentanglement in \acp{vae}. In Section \ref{sec:methodology} we cover background theory in order to provide a baseline from which to build our proposal. In Section \ref{sec:Experiments} we present the results from our quantitative and qualitative evaluations of Gated-VAE. Finally, a conclusion summarises the results and their significance.

\begin{figure}
\centering
\includegraphics{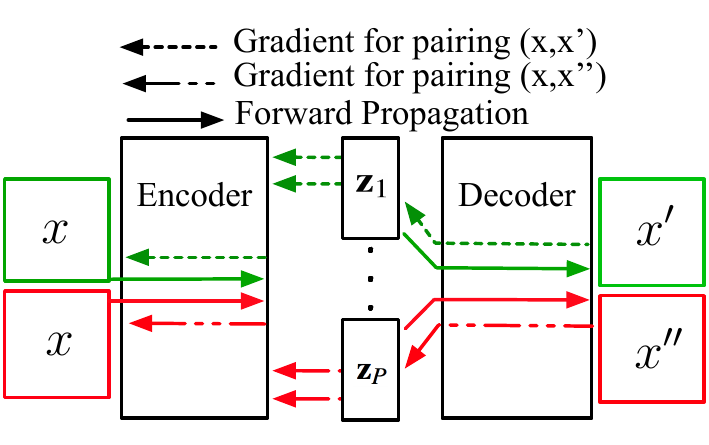}
\caption{Gated-VAE training principle. Images are paired according to shared subsets of latent factors, where the subsets are derived from any available supervision. A forward pass is made through the whole network, but gradients are backpropagated through a specific latent partition $\boldsymbol{z}_i$ where $i=[1...P]$ according to the input/target image pairing. }
\label{fig:simplediagram}
\end{figure}

\section{BACKGROUND}
\label{sec:background}
\acp{vae} are a type of generative model, and their performance for certain tasks is commonly compared against alternative generative models such as the Generative Adversarial Network (GAN) \cite{goodfellow2}. GANs have been shown to provide superior generative quality, but \acp{vae} have a number of advantages which include outlier robustness, improved training stability and interpretable, disentangled representations \cite{dai3}. Disentangled representations are generally conceived to be representations in which each element relates to an independent (and usually semantically meaningful) generative factor\cite{dai3, higgins}. Achieving a disentangled representation is suggested to aid in downstream tasks \cite{higgins}, however, there is some debate \cite{locatello, dai3} as to whether disentanglement helps \textit{per se}, or whether it is the \textit{informativeness} of the latent space that primarily determines the utility of the embedding - i.e. whether the representation fundamentally captures variation in the underlying factors. Disentanglement is essential if a self-contained subspace in the full latent embedding/representation needs to be extracted or masked for downstream purposes. For example, consider the task of unsupervised facial expression representation. In the wild, most images of faces will contain some variation in head-pose (i.e. on-axis frontal images of faces would represent the exception, not the norm). In this application it may therefore be useful to derive a representation of facial expression independent of head-pose. If head-pose could be reliably disentangled from facial expression, downstream tasks that depend on facial expression could be fed with expression representations invariant to head-pose. However, \acp{vae} do not currently disentangle with either predictability or consistency \cite{locatello}.

In recent years, various attempts have been made to encourage disentanglement with \acp{vae}, including increasing the emphasis on reducing the distance between the posterior and the prior \cite{higgins}, utilisation of alternative objective functions \cite{infovae, kumar2, kim4, chen5}, Gaussian mixtures \cite{gaussianvaes}, use of alternative prior distributions \cite{tomczak}, and cascaded \ac{vae} models \cite{dai3}. However, results from a review of 12,000 current implementations \cite{locatello} indicate that there is almost as much influence from random initialisation as there is from hyperparameter selection and objective functions. Indeed, consistent disentanglement has recently been demonstrated to be impossible without inductive bias and subjective validation \cite{locatello, dai3}. In other words, the typical evaluation of the inferred latent space is subjectively compared (e.g. using reconstructions of latent traversals) against our prior expectations / domain knowledge concerning which attributes should be disentangled for a given problem. Our work is concerned with incorporation of any available supervision or domain knowledge into the training procedure in order to encourage disentanglement.

Beyond subjective interpretation, there is not yet a consensus on the best way to quantitatively measure disentanglement, although various proposals have been made. These include Separated Attribute Predictability \cite{kumar2}, Mutual Information Gap \cite{chen4}, FactorVAE metric \cite{kim4}, Modularity \cite{ridgeway}, the $\beta$-\ac{vae} metric \cite{higgins} and the later relative of the $\beta$-\ac{vae} metric \cite{eastwood}. Whether any of these metrics measure disentanglement as it is generally conceived is unclear \cite{locatello}. The metrics proposed by \cite{eastwood} have been chosen for our evaluation and represent one of the most recent attempts to measure disentanglement and distinguish between disentanglement and informativeness, as well as providing an estimation of completeness (these terms are described in more detail in Section \ref{sec:syntheticevaluation}). We utilise these metrics to contribute insight into the relationship between informativeness and disentanglement.

%%%%%%%%%%%%%%%%%%%%%%%%%%%%%%%%%%%%%%%%%%%%%%%%%%%%%%%%%%%%%%%%%%%%%%%%%%%%%%%%%%

\section{METHODOLOGY}
\label{sec:methodology}
This section begins with an overview of \ac{vae} theory before a presentation of the currently proposed formulation.
% Put brief intro to this section

\subsection{Variational AutoEncoders - Background Theory}

The reader is directed to \cite{doersch, kingma, rezende2} for a more detailed introduction to \acp{vae}. In essence, and following the process for variational inference for a distribution of latent variables, we start by sampling from a latent distribution $\mathbf{z} \sim p(\mathbf{z})$ and generate dataset $X$ of images $\mathbf{x} \in \mathbb{R}^N$ with observational distribution $p_\theta(\mathbf{x}|\mathbf{z})$ such that we may derive an inferred posterior for the latent distribution as $q_\phi(\mathbf{z}|\mathbf{x})$ that approximates the true conditional latent distribution $p_\theta(\mathbf{z}|\mathbf{x})$. Both $q_\phi(\mathbf{z}|\mathbf{x})$ and $p_\theta(\mathbf{x}|\mathbf{z})$ are parameterised by neural network encoder and decoder parameters $\phi$ and $\theta$ respectively \cite{tshannen, doersch,infovae}. The traditional approach \cite{kingma,kumar2} involves maximisation of the Evidence Lower BOund (ELBO):

\begin{equation}
\begin{split}
 \max _{\theta, \phi}\mathbb{E}_{\mathbf{x}}\left[ \mathcal{L}_{\mathrm{ELBO}}(x)\right] =
\\ 
\max _{\theta, \phi} \mathbb{E}_{\mathbf{x}}\left[\mathbb{E}_{\mathbf{z} \sim q_{\phi}(\mathbf{z} | \mathbf{x})}\left[\log p_{\theta}(\mathbf{x} | \mathbf{z})\right]- \beta \mathrm{KL}\left(q_{\phi}(\mathbf{z} | \mathbf{x}) \| p(\mathbf{z})\right)\right]
\end{split}
\label{eq:ELBO}
\end{equation}

The first term on the RHS of Eq. \ref{eq:ELBO} encourages reconstruction accuracy, and the Kullback-Liebler divergence term (weighted by parameter $\beta$ \cite{higgins}) acts as a regulariser, penalising approximations for $q_\phi(\mathbf{z}|\mathbf{x})$ that do not resemble the prior. The objective is therefore to maximise the marginal log-likelihood of $\mathbf{x}$ over the latent distribution $\mathbf{z}$ \cite{higgins}, which is assumed to be Gaussian with identity covariance $\mathbf{z} \sim \mathcal{N}(0, \mathbf{I})$. The Gaussian assumption means that Eq. \ref{eq:ELBO} may be written using an analytical reduction of the KL divergence term \cite{kumar2}:
\begin{equation}
\label{eq:ELBO2}
\begin{split}
\max_{\theta, \phi}\mathbb{E}_{\mathbf{x}} 
    \left[ 
        \mathcal{L}_{\mathrm{ELBO}}(x) 
    \right] 
    = 
    \max_{\theta, \phi} \mathbb{E}_{\mathbf{x}}
    \biggl[ 
        \mathbb{E}_{\mathbf{z} \sim q_{\phi}(\mathbf{z} | \mathbf{x})}
        \left[ 
            \log p_{\theta}(\mathbf{x} | \mathbf{z}) 
        \right] - \\ 
        \frac{\beta}{2}
        \bigl(
            \sum_{i}
            \left(
                \left[
                    \Sigma_{\phi}(\mathbf{x})
                \right]_{ii} - \ln 
                \left[ 
                    \boldsymbol{\Sigma}_{\phi}(\mathbf{x})
                \right]_{ii}
            \right) + 
            \left\|
                \boldsymbol{\mu}_{\phi}(\mathbf{x})
            \right\|_{2}^{2}
        \bigr)
    \biggr] 
\end{split}
\end{equation}
In Eq. \ref{eq:ELBO2} the $\left[\Sigma_{\phi}(\mathbf{x})\right]_{i i}$ indicates the diagonal covariance, and $\boldsymbol{\mu}_{\phi}(\mathbf{x})$ is the mean. Both the mean and covariance are learned by the network encoder and parameterise a multivariate Gaussian that forms the inferred latent distribution $q_\phi(\mathbf{z}|\mathbf{x})$. The decoder network samples from  $\mathbf{z} \sim q_\phi(\mathbf{z}|\mathbf{x})$ using the reparameterisation trick \cite{doersch} such that $\mathbf{z} = \mu_\phi(\mathbf{x}) + \epsilon\sqrt{\Sigma_\phi(\mathbf{x})}$ where $\epsilon = \mathcal{N}(0, \mathbf{I})$. One interpretation of disentanglement posits that it is achieved if $q_\phi (\mathbf{z}) = \int{ q_\phi(\mathbf{z}|\mathbf{x})p(\mathbf{x})d\mathbf{x}} = \prod_i q_i(\mathbf{z}_i)$ \cite{kumar2}.

 During VAE training, the $i$\textsuperscript{th} reconstructed example from the output of the network decoder $\hat{x}_i$ is typically compared against the $i$\textsuperscript{th} input example $x_i$, which is therefore also used as the target, and the marginal log likelihood is maximised by minimising a loss (e.g. binary cross-entropy) between the reconstruction and the target. Over the course of training, the VAE thereby learns decoder parameters that produce the best reconstruction, conditioned on the latent embedding of the input image. In order for the \ac{vae} to infer the generative distribution $\mathbf{z}$, the ground truth latent factors must both vary between examples in the dataset, \textit{and} take on the same value in both input and target pairs. Letting each image in $X$  have $K$ independent generative ground truth factors $\mathbf{v} \in \mathbb{R}^K$ i.e. $\{\mathbf{v}_k \}^K_{k=1}$ which we model using a latent generative distribution $\mathbf{z} \in \mathbb{R}^M$ such that $M \geq K$. Given that the input image $x_{q}$ is usually equal to the target image $x_r$ (as is usual for auto-encoding), the values of all ground truth generative factors in the input image are therefore the same as the values in the target image i.e. $v_{k, x_{q}} = v_{k, x_{r}} , k=1...K $, where $v_{k, x_{q}}$ is the ground truth factor $k$ for the input image, and $v_{k, x_{r}}$ is the same ground truth factor $k$ for the target image.
 
Readers are referred to \cite{tshannen} for a review of the various modifications proposed to the objective in Eq. \ref{eq:ELBO} that encourage disentanglement and minimise reconstruction cost. However, as mentioned, there is some evidence to suggest that the random initialisation of the network has almost as much impact as the network architecture and objective functions have on the ability of the network to disentangle the latent space \cite{locatello}. Furthermore, VAE disentanglement is often evaluated against prior subjective expectations, and therefore, by definition, we are utilising domain knowledge to evaluate the success of an unsupervised method. For instance, Higgins et al. \cite{higgins} generate reconstructions of traversals/interpolations over the latent space in order to visually ascertain which of the latent space dimensions correspond e.g. with shape or rotation. Furthermore, it is often the case that some form of weak supervision is available for any given task. For example, if we have frames from a sequence, it is likely that the appearance and identity of individuals is unlikely to change between subsequent frames within that sequence. In more extreme cases, full supervision for all generative factors may be available and utilised to achieve disentanglement in an entirely prescribed way (e.g. see \cite{kulkarni}). We propose the Gated-VAE and thereby provides a means for domain knowledge to be employed at the training stage in order to encourage disentanglement in a novel, weakly-supervised way. 

\subsection{Gated Variational AutoEncoders - Formulation}
\label{sec:methodologygated}

Often, weak supervision is available in some form (e.g. data may be clustered, or have weak labels). If any supervision is available then it should be incorporated into training in order to aid disentanglement. A Gated-VAE provides a means to incorporate available supervision into existing VAE models. The intuition behind the Gated-VAE is that input and target images can be paired according to shared factors, and that the network should learn to recognise and learn what is common between these pairs. In other words, by pairing the input image with a target image that shares specific latent factors, the network can be encouraged to disentangle these shared factors. Such pairing may be possible when weak supervision (e.g. clustering) is available. If the supervision is available then it ought to be incorporated where possible. Backpropagation of error can then be directed through specified partitions of the latent space such that different partitions are disentangled and each contains information relating to the shared factors. The approach is deemed to be weakly supervised because the input and target images need to be paired according to some prior knowledge or labels. Weak supervision is generally used to describe the scenario whereby labels are available but the labels only relate to a limited number of factors (e.g. labels may be fully supervised and describe head-pose in terms of roll, pitch and yaw, or be weakly supervised and simply indicate that two images simply share the same head-pose) \cite{szabo1}. Weak supervision should not be confused with semi-supervision whereby fully informative labelling is available but only for a subset of the data \cite{kingma3}. 

More concretely, we can define a subset of latent factors $\mathbf{s} \subset \mathbf{v}$ such that $\mathbf{s} \in \mathbb{R}^L$ where $L < K \leq M$. We can partition the latent space and train each partition by using input/target pairs where $x_{q} \neq x_{r}$ but where $s_{l, x_{q}} = s_{l, x_{r}}, \forall l$. The partition will learn factors $s_l$ but not the factors in $\mathbf{v}$ that are not in $\mathbf{s}$. Designing the input/target pairs in such a way requires domain knowledge, and therefore deviates from unsupervised training to semi-supervised training. 
 
 Starting with a \ac{vae} with an $M$-dimensional latent space, we split the latent space into $P$ partitions (that need not be equal in size) such that $q_{\phi}(\mathbf{z}|\mathbf{x})$ is parameterised as follows:

\begin{equation}
\begin{split}
\mathbf{z} = [\mathbf{z}_1, \mathbf{z}_2, ..., \mathbf{z}_P]  = \\ \left[\left(\mu_{\phi, 1}(\mathbf{x}) + \epsilon\sqrt{\Sigma_{\phi, 1}(\mathbf{x})}\right), ..., \left( \mu_{\phi, P}(\mathbf{x}) + \epsilon\sqrt{\Sigma_{\phi, P}(\mathbf{x})}\right)\right] 
\end{split}
\end{equation}

During forward propagation, all partitions of the latent space are used and concatenated together. Similarly, the computation of the KL divergence is also taken over the entire latent space either by concatenating the partitions and computing the loss, or by computing the loss over each partition and concatenating the loss. However, during back-propagation, the gradient is gated according to the input/target image pairing. 

An example of two training iterations with $P$ partitions is depicted in Figure \ref{fig:simplediagram}. In the first training iteration, Image $x$ is paired with target Image $x'$ such that $x \neq x'$ but that $x$ and $x'$  have equal ground truth latent factors for a subset of all ground truth factors, and gradients are backpropagated from end-to-end but only through a specific partition of the inferred latent space. For the second training iteration Image $x$ is paired with target image $x''$ such that  $x$ and  $x''$ also have ground truth factors that are equal for only a subset of all ground truth factors (but a different and possibly overlapping subset to the subset shared between  $x$ and  $x'$), and gradients are backpropagated from end-to-end but through a different partition of the latent space. Note that the decoder has access to the entire latent space on the forward passes to generate the reconstruction. If the pairing of images is consistent according to the desired partitioning across the entire dataset, then the partitions will contain different factors. Even if disentanglement has not occurred \textit{within} each partition, disentanglement will occur \textit{between} the partitions. 

In cases where reconstruction quality is desired then the training process can be split into two - firstly to prioritise disentanglement using the Gated-VAE method, and secondly by fixing the weights of the encoder and latent space, and fine-tuning the decoder by continuing its training to maximise reconstruction quality using the traditional method for training \acp{vae} (where input image $x_{q}$ is identical to target image $x_{r}$).

\begin{figure*}[!ht]
\centering
  \includegraphics[trim={0.2cm 0.2cm 0.2cm 0.2cm},clip,width=1\linewidth]{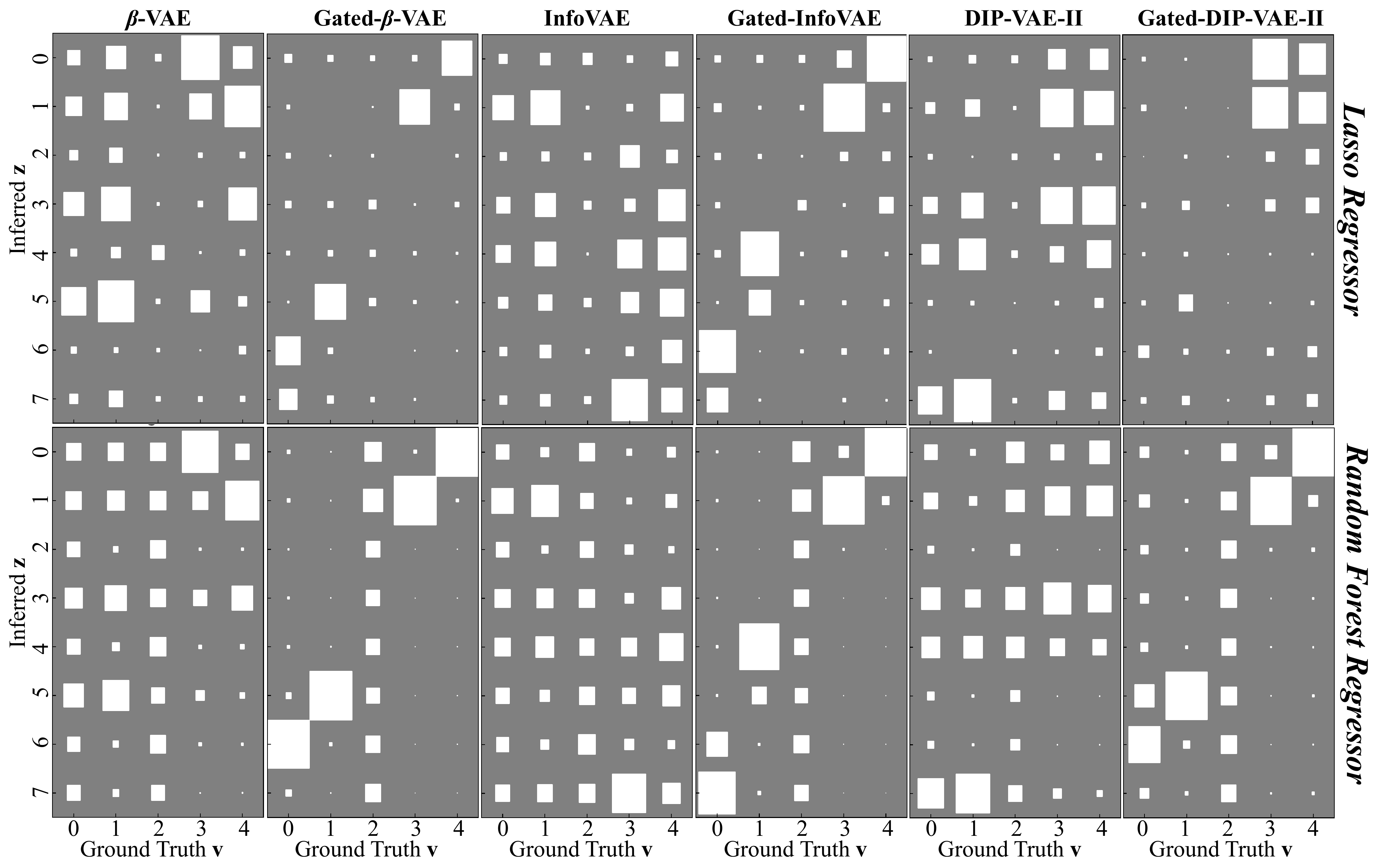}
  \caption{Example Hinton diagrams for each of the gated and un-gated models depicting the relative importance of each inferred latent dimension $\mathbf{z}$ for predicting the ground truth generative factors $\mathbf{v}$. $v_0 =$ shape, $v_1$ = size, $v_2=$ rotation, $v_3=$ x-position and $v_4=$ y-position.}
  \label{fig:hintons}
\end{figure*}%

\section{EXPERIMENTS}
\label{sec:Experiments}
The experiments demonstrate that the weakly-supervised Gated-VAE can be used to adapt existing VAE models in order to improve disentanglement. The method is first tested on synthetic data, and then on a dataset of faces.

\subsection{Synthetic Data}
 We begin by demonstrating the quantitative performance improvement achieved by applying gating to three non-convolutional implementations of existing VAE models: $\beta$-VAE \cite{higgins}, InfoVAE \cite{infovae} and DIP-VAE-II \cite{kumar2}. The experiments were undertaken 10 times in order to acquire averages and standard deviations for the quantitative metrics. 

\begin{figure}[H]
\centering
  \includegraphics[width=0.5\linewidth]{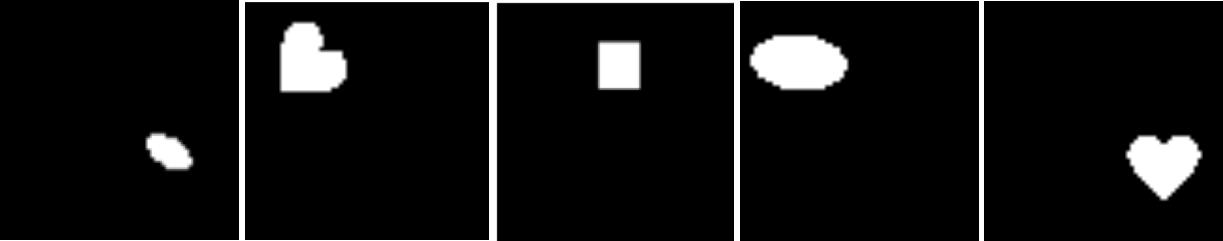}
 \caption{Samples from the dSprites \cite{dsprites} dataset, varying in x-position, y-position, rotation, size and shape.}
  \label{fig:dsprites:examples}
\end{figure}%

The dSprites \cite{dsprites} dataset was used for initial experimentation and comprises 737280 (64x64) images of white shapes on a black background that vary over only five generative factors: $v_0 =$ shape (square, ellipse, heart), $v_1$ = size (6 sizes linearly spaced), $v_2=$ rotation (40 values over $2\pi$), $v_3=$ x-position (32 values) and $v_4=$ y-position (32 values). This dataset was chosen as it is a common baseline used to test VAEs \cite{higgins, disentanglement, locatello}. Samples from the dSprites dataset can be seen in Figure~\ref{fig:dsprites:examples}.

Input/target image pairs are chosen according to equal manifestations of generative factors. For example, in order to train the partition that is intended to learn the size of the shape in the image, a random batch of images is chosen for the input images, and the corresponding target batch is chosen such that the size of the shape in each of the target images is the same as the size of the shape in each of the paired input images. The same batching process can be applied to pair images with the same x/y position, or the same shape etc. One partition was used to represent both x and y position dimensions in order to demonstrate whether these two dimensions can be disentangled from each other within a single partition (i.e. an input image with a certain x/y position was paired with a target image with the same x/y position such that two generative factors were shared and only one partition was gated). Note that, in the case of the dSprites dataset, full supervision is available and is being used to identify the pairs. However, the labels are not provided explicitly to the network by virtue of the input/target pairing process of the Gated-VAE.
\\
\subsubsection{Evaluation - Synthetic Data}
\label{sec:syntheticevaluation}
The disentanglement, completeness and informativeness metrics from \cite{eastwood} are used for quantitative evaluation of the Gated-VAE. These metrics build on the $\beta$-VAE metric proposed in \cite{higgins} and are derived using linear Lasso and non-linear Random-Forest (RF) regressors that predict the ground truth factors using the latent embeddings of a test dataset. The regressors provide a matrix of relative importance, representing the importance of each inferred latent dimension for predicting each of the ground truth generative factors. These matrices are used to generate Hinton diagrams \cite{hintondiagrams} for visualisation purposes. 

According to the definitions in \cite{eastwood,bengio1, higgins}, \textit{disentanglement} describes the degree to which each inferred factor in $\mathbf{z}$ independently predicts a corresponding ground truth factor. Concretely, and as defined in \cite{eastwood}, disentanglement of inferred factor $z_i = \mu_{\phi,i}(\mathbf{x})$ is calculated as $D_i = (1-H_K(P_i))$ where $H_K(P_i) = -\sum_{k=0}^{K-1}P_{ik}\log_KP_{ik}$ is the entropy and $P_{ij} = R_{ij}/ \sum_{k=0}^{K-1}R_{ik}$ is the probability that inferred factor $z_i$ is used by a classifier or regressor to predict ground truth factor $v_j$ (which is a modified form to \cite{eastwood} to be consistent with terms in this paper). A weighting is calculated using $\rho_i = \sum_jR_{ij} / \sum{ij}R_{ij} $ so that the average disentanglement is weighted using $\sum_i \rho_iD_i$. Using these definitions means that a disentanglement score of approximately 0 corresponds with an inferred variable that does not contribute any predictive power. 

\textit{Completeness} \cite{eastwood} is complementary to disentanglement in that it is calculated using the same relationships for disentanglement as set out above but for each of the $M$ inferred latent dimensions, rather than the $K$ generative factors. A score of 1 for a particular generative factor $v_k$ means that this generative factor is predicted by a single inferred latent dimension, and a score of 0 means the generative factor is predicted equally by all inferred latent dimensions.

 \textit{Informativeness} \cite{eastwood} describes whether the inferred latent representation is useful in predicting ground truth factors. In other words, it tells us whether the latent space contains useful information about the generative factors. It is quantified using the average regressor prediction error such that a lower prediction error corresponds with a higher informativeness and therefore a lower value is desirable. We used the normalised root mean squared error for the experimental results. 
 
 Given these definitions for informativeness and disentanglement, informativeness becomes distinct from disentanglement, as an inferred latent representation may be highly informative without necessarily being disentangled. The characteristics of the metrics motivate the use of the RF (i.e. non-linear) regressor, because learned embeddings do not need to be linear with respect to their corresponding ground truth factors \cite{eastwood}. For example, any rotational factors which `wrap' around $2\pi$ may be embedded sinusoidally. Similarly, learned embeddings capturing more than one generative factor simultaneously may be non-linearly informative, even though they are not disentangled. The results for both the Lasso regressor and RF regressors are presented in Section \ref{sec:syntheticresults}. The Lasso regularisation parameter, $\alpha = 0.02$ for all runs and the RF utilised 10 estimators each with a maximum depth of 12.
\\
\subsubsection{Models}
Gating will be applied to three recently proposed variants of VAEs: $\beta$-VAE \cite{higgins} (which increases the pressure on the KL-divergence loss), InfoVAE \cite{infovae} (which minimises maximum mean discrepancy) and DIP-VAE-II \cite{kumar2} (which minimises the 2nd central moment of the latent space). Readers are referred to the original papers for a more detailed description of these models. In terms of parameter values, for $\beta$-VAE, $\beta = 4$ (as suggested by \cite{higgins}), for DIP-VAE-II, $\lambda_{od} = \lambda_d = 250$, and for InfoVAE $\lambda_v=500$ where all $\lambda_{\_}$ parameters represent a weight on the respective component(s) of the models' objective functions. 

The latent space $\mathbf{z}$ for all models had dimensionality $M=8$ and was split into $P=4$ partitions of equal size. $M=8$ was chosen so that each partition could represent at most 2 generative factors, where there are 5 generative factors in total.  The architecture of the basic network encoder comprises 2 fully connected layers with batch normalization and ReLU activations, and the decoder comprises 3 fully connected layers with batch normalization, ReLU activations for the first two layers and a sigmoid activation at the output. Modifications to the loss functions are made to adapt each network for $\beta$-VAE, InfoVAE and DIP-VAE-II. For all models, an Adam \cite{adam} optimiser was used with a learning rate of 0.0001, and the network was trained for 50 epochs with $N=128$. \\ 

\begin{table*}
\centering
\caption{Averages weighted by predictor importance with standard deviations for disentanglement, completeness and informativeness with and without gating for $\beta$-VAE \cite{higgins}, InfoVAE \cite{infovae} and DIP-VAE-II \cite{kumar2} over 10 runs. Note that a lower score is desired for informativeness.}
\label{tab:results}
\begin{tabular}{llllll}
\hline
\textbf{Regressor} & \textbf{Model} & \textbf{    } & \textbf{Disent.} & \textbf{Complete.} & \textbf{(Un)Inform.} \\ \hline
\multirow{6}{*}{Lasso} 
% Lasso:\beta-VAE results
& \multirow{2}{*}{$\beta$-VAE}
& -- & \multicolumn{1}{c}{0.237$\pm0.039$} & \multicolumn{1}{c}{0.276$\pm0.048$} &  \multicolumn{1}{c}{0.690$\pm0.019$} \\
& & \textbf{Gated} & \multicolumn{1}{c}{\textbf{0.609}$\pm0.136$}  & \multicolumn{1}{c}{\textbf{0.478}$\pm0.104$} & \multicolumn{1}{c}{\textbf{0.432}$\pm0.087$} \\
% Lasso:InfoVAE results
& \multirow{2}{*}{InfoVAE} 
& -- & \multicolumn{1}{c}{0.240$\pm0.034$} & \multicolumn{1}{c}{0.156$\pm0.020$} &  \multicolumn{1}{c}{0.772$\pm0.013$} \\
& & \textbf{Gated} & \multicolumn{1}{c}{\textbf{0.647}$\pm0.091$}  & \multicolumn{1}{c}{\textbf{0.495}$\pm0.070$} & \multicolumn{1}{c}{\textbf{0.481}$\pm0.062$} \\
% Lasso:DIP-VAE-II results
& \multirow{2}{*}{DIP-VAE-II}
& -- & \multicolumn{1}{c}{0.316$\pm0.113$} & \multicolumn{1}{c}{0.346$\pm0.091$} &  \multicolumn{1}{c}{0.653$\pm0.033$} \\
& & \textbf{Gated} & \multicolumn{1}{c}{\textbf{0.487}$\pm0.093$}  & \multicolumn{1}{c}{0.337$\pm0.081$} & \multicolumn{1}{c}{0.604$\pm0.086$} \\
\hline
\multirow{6}{*}{\begin{tabular}[c]{@{}l@{}}Random\\ Forest\end{tabular}} 
% RandomForest: \beta-VAE results
& \multirow{2}{*}{$\beta$-VAE}
& -- & \multicolumn{1}{c}{0.172$\pm0.033$} & \multicolumn{1}{c}{0.237$\pm0.032$} & \multicolumn{1}{c}{0.460$\pm0.007$} \\
& & \textbf{Gated} & \multicolumn{1}{c}{\textbf{0.631}$\pm0.113$} &  \multicolumn{1}{c}{\textbf{0.667}$\pm0.093$} &\multicolumn{1}{c}{\textbf{0.258}$\pm0.046$} \\
% RandomForest: InfoVAE results
& \multirow{2}{*}{InfoVAE}  
& -- & \multicolumn{1}{c}{0.113$\pm0.030$} & \multicolumn{1}{c}{0.142$\pm0.026$} & \multicolumn{1}{c}{0.483$\pm0.004$} \\
& & \textbf{Gated} & \multicolumn{1}{c}{\textbf{0.632}$\pm0.047$} &  \multicolumn{1}{c}{\textbf{0.646}$\pm0.032$} &\multicolumn{1}{c}{\textbf{0.243}$\pm0.012$} \\
% RandomForest: DIP-VAE-II results
& \multirow{2}{*}{DIP-VAE-II}
& -- & \multicolumn{1}{c}{0.197$\pm0.097$} & \multicolumn{1}{c}{0.346$\pm0.095$} & \multicolumn{1}{c}{0.456$\pm0.008$} \\
& & \textbf{Gated} & \multicolumn{1}{c}{\textbf{0.486}$\pm0.076$} &  \multicolumn{1}{c}{\textbf{0.527}$\pm0.077$} &\multicolumn{1}{c}{\textbf{0.394}$\pm0.043$} \\\hline
\end{tabular}
\end{table*}

\subsubsection{Results - Synthetic Data}
\label{sec:syntheticresults}

Figure \ref{fig:hintons} shows example Hinton diagrams for the relative importance of each of the inferred latent dimensions $\mathbf{z}$ for predicting the ground truth factors $\mathbf{v}$, with and without our proposed gating, as well as using Lasso and RF regression. In an ideal result, the Hinton diagram would contain five distinct squares, with only one square per column and one per row. Figure \ref{fig:hintons} demonstrates consistent allocation of four out of five factors to their intended partitions in the latent space. Unfortunately, rotation $(v_2)$ was not well encoded by \textit{either} gated or un-gated models. The informativeness (i.e. NRMSE) of rotation for the  gated and un-gated $\beta$-VAE models were 0.985$\pm 0.004$ and 0.964$\pm 0.004$ respectively. The results are similar for the other two models: RF rotation informativeness for gated and un-gated DIP-VAE-II are 0.991$\pm0.003$ and 0.983$\pm0.008$ respectively; and 0.985$\pm0.004$ and 0.967$\pm0.003$ for gated and un-gated InfoVAE (see supplementary material for complete results). The NRMSE informativeness results with the Lasso regressor are even closer to 1 for both gated and un-gated models.

The results also indicate that the x/y position factors were disentangled \textit{within} a single latent partition. The examples shown in Figure \ref{fig:hintons} suggest that this was achieved more successfully than the un-gated equivalents. This may be because the additional supervision afforded by the Gated-VAE modification reduces the number of dimensions the VAE is disentangling at any one time.

Table \ref{tab:results} shows the results for disentanglement, completeness and informativeness for the three gated and un-gated models averaged over 10 runs using $N=128$ for the Lasso and RF regressors. In all cases disentanglement was increased with the use of the Gated-VAE as expected. Interestingly, the informativeness of the latent space was also increased, suggesting that the Gated-VAE improves the usefulness of the learned embedding by embedding more information than the un-gated equivalents. Furthermore, in all cases apart from those for DIP-VAE-II with Lasso, gating achieved better performance in completeness than the un-gated equivalent. For DIP-VAE-II vs. Gated-DIP-VAE-II with Lasso, disentanglement was significantly improved in spite of comparable informativeness. This suggests that a space may be disentangled without being informative, or vice versa.

The consistent improvement of RF over Lasso suggests the latent representation contained information about the factors that was encoded in such a way that the non-linear RF regressor was able to fit the relationships between $\mathbf{z}$ and $\mathbf{v}$ but the Lasso regressor could not. The results demonstrate how a latent representation may be informative even if the factors are disentangled by any arbitrary degree. The improvement of the Gated-DIP-VAE-II over DIP-VAE-II was less pronounced than for the other models. This is likely to be due to the relatively poor performance of DIP-VAE-II compared with the other models, although based on previous indications of the DIP-VAE-II's performance \cite{kumar2}, the low performance itself may be due to the particular hyperparameters used in the current experiments, and further work should involve hyperparameter optimisation of the DIP-VAE-II model.

\begin{figure}[!h]
\centering
  \includegraphics[trim={0 0 0 1.25cm},clip,width=0.7\linewidth]{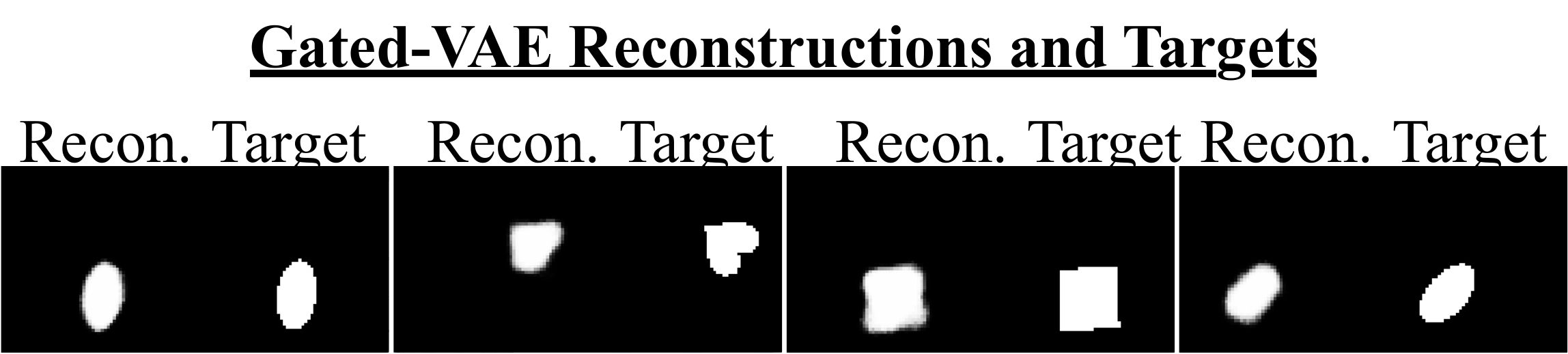}
  \caption{Reconstructions and corresponding targets for the Gated-$\beta$-VAE network with a fine-tuned decoder that demonstrate how location, size, rotation, and shape were all recovered.}
  \label{fig:finetuned}
\end{figure}%

Finally, Figure \ref{fig:finetuned} depicts Gated-$\beta$-VAE reconstructions. The images demonstrate how, despite the poor regressor metrics, all factors (including rotation) were nonetheless encoded. In order to `fine-tune' the network, all encoder parameters for the trained Gated-VAE were fixed, and the network decoder was then trained for 1 epoch according to the typical VAE training procedure (where the input and target images are identical).

\subsection{Face Data}

In order to demonstrate the effectiveness of Gated-VAE on more complex data, the CelebA \cite{celebA} dataset is used. The CelebA dataset comprises 202,599 faces of 10,177 different individuals. The dataset was converted to greyscale to expedite training. OpenFace 2.0 \cite{openface2} was used both to align the faces and also to generate labels for head-pose (pitch and yaw\footnote{No roll labels were used because OpenFace 2.0 aligns faces according to horizontal eye position and thereby removes variation in the roll dimension.}) and facial expression (Facial Action Units - FACS), and thereby provide a source of weak supervision with which to train the Gated-VAE. The images were then clustered using K-Means, yielding 2,500 clusters for head-pose and 4,000 for facial expression. Such a clustering method is clearly not optimal if the goal is to achieve accurate labels and ideal image pairings. However, accurate supervision may rarely be available in real-world applications, and so this method is actually well-suited as a demonstration of how to encourage disentanglement when only noisy/weak supervision is available. \\

\subsubsection{Evaluation - Face Data}
\label{sec:facesevaluation}
As only weak labels (as opposed to ground truth) are available, we are unable to undertake the quantitative evaluation of performance that was used with the synthetic data. Instead, a qualitative evaluation is performed by generating reconstructions of latent traversals. Such a method is common in the VAE literature \cite{higgins, disentanglement} and is recommended as a means of model diagnosis \cite{theis}. \\

\begin{figure*}[!ht]
\centering
  \includegraphics[width=0.7\linewidth]{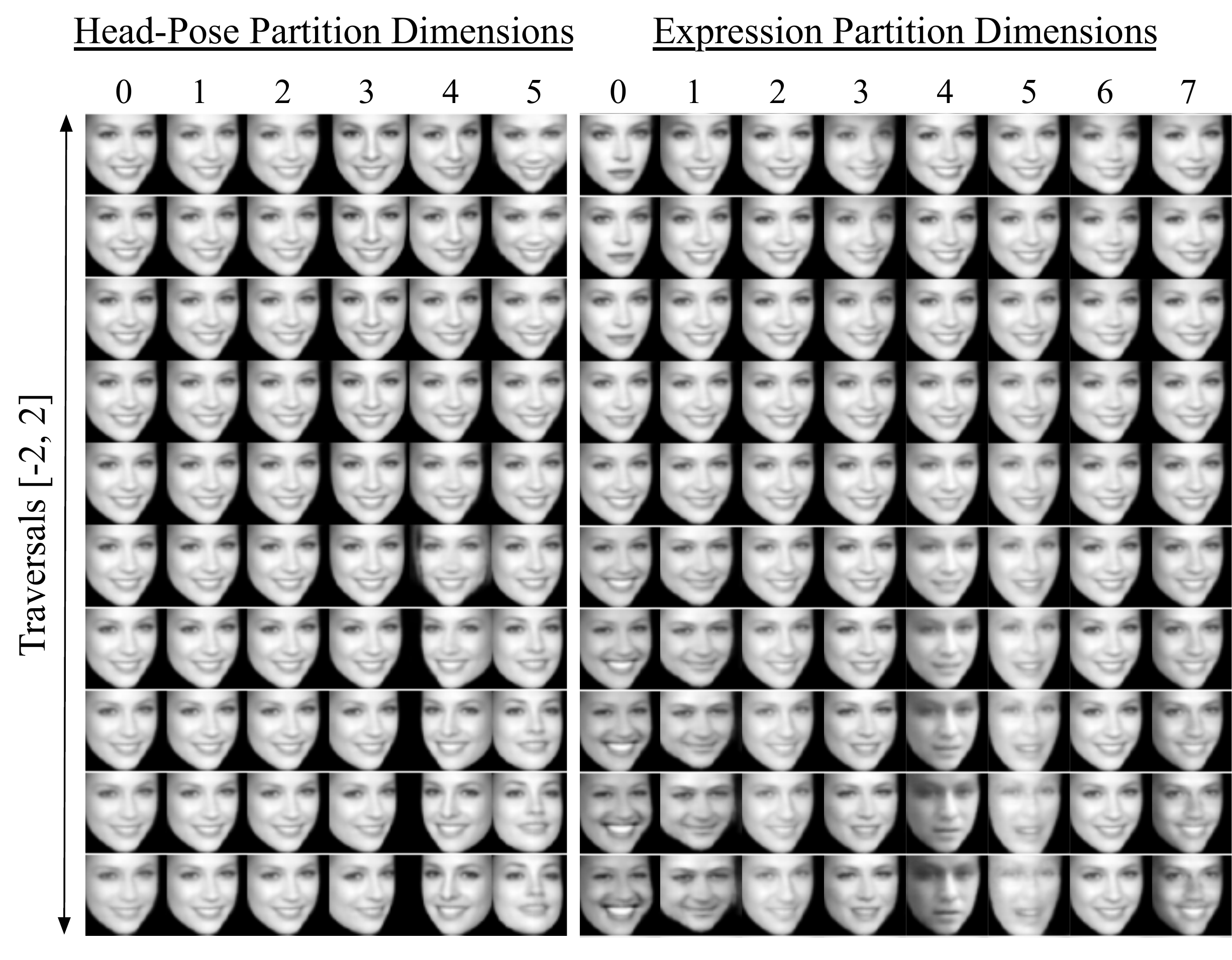}
  \caption{Reconstructions of traversals (between $\pm2$) of the two partitions (head-pose and expression) of the latent space for Gated-VAE. Only the 8 most active dimensions for the expression partition are shown.}
  \label{fig:traversals}
\end{figure*}%

\subsubsection{Model}
\label{sec:facesmodels}
Gating will be applied to a convolutional implementation of the vanilla VAE network VAE which has a weight $\beta>1$ on the KL divergence term in the objective function). This is because increasing the $\beta$ term has been shown to increase low-pass filtering characteristics (i.e. removing detail) as a side-effect of disentanglement \cite{disentanglement}. Given also that $\beta$-VAE has been shown \cite{locatello} not to disentangle latent factors in a way that is consistent (at least not with our subjective expectations) we leave disentanglement entirely to the encouragement afforded by the gating with weak-supervision. The latent space comprised two partitions of 6 (for head-pose) and 18 (for expression) dimensions. \\

\subsubsection{Results - Face Data}
\label{sec:facesresults}
Reconstructions of latent space traversals are shown in Figure \ref{fig:traversals}. These traversals are generated by sampling and encoding a random image from the dataset, and then interpolating along each dimension step-wise between $\pm 2$. It can be seen that head-pose has been disentangled from the rest of the generative factors, despite some degree of spill from facial expression appearing in dimension 5 of the head-pose partition. Similarly, the expression partition does not appear to contain any head-pose information, although it is interesting to note that appearance (e.g. skin colour and gender) has been encoded in this partition, despite not having been provided with supervision for this subset of factors. It can be seen that dimensions 0-2 for the head-pose partition were not used, which is as expected given that pitch and yaw can be optimally encoded with only two dimensions.

\section{CONCLUSIONS AND FUTURE WORKS}
We have presented a weakly-supervised modification to the training process for \acp{vae} which involves the partitioning of the latent space and restriction or `gating' of gradients during backpropagation through the partitions where the gating depends on the chosen input/target image pairings. The Gated-VAE modification allows for domain knowledge to be incorporated into the training process, and can be applied to existing VAE models. The experiments compared the performance of $\beta$-VAE, DIP-VAE-II and InfoVAE with and without gating using the evaluation metrics proposed by \cite{eastwood}, and a qualitative demonstration was presented in the form of latent traversals with the CelebA \cite{celebA} dataset.

Gated versions outperformed the un-gated equivalent models in disentanglement, completeness and informativeness. The relationship between disentanglement and informativeness was illustrated, in that a disentangled latent space did not necessarily imply informativeness. The Hinton diagrams demonstrate how gating consistently allocates the relevant latent variables to a partition in the latent space thereby achieving reliable disentanglement between partitions as well as producing a more informative (and therefore more useful) latent space than the un-gated equivalent. The Hinton diagrams and individual factor results for informativeness illustrate how the rotation factor was not well encoded by either gated or un-gated models, despite being sufficiently encoded to facilitate correct reconstruction (Figure \ref{fig:finetuned}). This may be due to the fact that the rotation factor is lower in its salience with respect to the pixel-wise cross-entropy loss as compared with size, shape, and position. The un-gated models also had negligible informativeness for rotation.  The demonstration of Gated-VAE on the CelebA dataset demonstrated that, even with noisy, weak supervision (from clustered OpenFace 2.0 output), compelling disentanglement between head-pose and facial expression was nonetheless achieved. The Gated-VAE's consistent allocation of factors to intended partitions in the latent space provides a means to mask or extract informative partitions for the purposes of downstream tasks. Further work is recommended to establish the efficacy of Gated-VAE for other downstream tasks including human-computer interaction technologies and sign language translation.

\bibliographystyle{unsrt}  

\begin{thebibliography}{10}

\bibitem{infovae}
S.~Zhao, J.~Song, and S.~Ermon.
\newblock Info{VAE}: Balancing learning and inference in variational
  autoencoders.
\newblock {\em arXiv:1706.02262v3}, 2018.

\bibitem{tshannen}
M.~Tschannen, O.~Bachen, and M.~Lucic.
\newblock Recent advances in autoencoder-based representation learning.
\newblock {\em arXiv:1812.05069v1}, 2018.

\bibitem{sonderby}
C.~K. Sonderby, T.~Raiko, L.~Maaloe, S.~K. Sonderby, and O~Winther.
\newblock How to train deep variational autoencoders and probabilistic ladder
  networks.
\newblock {\em arXiv:1602.02282v1}, 2016.

\bibitem{higgins}
I.~Higgins, L.~Matthey, A.~Pal, C.~Burgess, X.~Glorot, M.~Botvinick,
  S.~Mohamed, and A.~Lerchner.
\newblock Beta-{VAE}: Learning basic visual concepts with a constrained
  variational framework.
\newblock {\em ICLR}, 2017.

\bibitem{kumar2}
A.~Kumar, P.~Sattigeri, and A.~Balakrishnan.
\newblock Variational inference of disentangled latent concepts from unlabeled
  observations.
\newblock {\em arXiv:1711.00848v3}, 2018.

\bibitem{bengio1}
Y.~Bengio, A.~Courville, and P.~Vincent.
\newblock Representation learning: A review and new perspectives.
\newblock {\em IEEE Transactions on pattern analysis and machine intelligence},
  2013.

\bibitem{dai3}
B.~Dai and D.~Wipf.
\newblock Diagnosing and enhancing {VAE} models.
\newblock {\em arXiv:1903.05789v1}, 2019.

\bibitem{locatello}
F.~Locatello, S.~Bauer, M.~Lucic, G.~Ratsch, S.~Gelly, B.~Scholkopf, and Bachem
  O.
\newblock Challenging common assumptions in the unsupervised learning of
  disentangled representations.
\newblock {\em arXiv:1811.12359v3}, 2019.

\bibitem{eastwood}
C.~Eastwood and C.~K.~I. Williams.
\newblock A framework for the quantitative evaluation of disentangled
  representations.
\newblock {\em ICLR Conference}, 2018.

\bibitem{goodfellow2}
I.~J. Goodfellow, J.~{Pouget-Abadie}, M.~Mirza, B.~Xu, D.~{Warde-Farley},
  S.~Ozair, A.~Courville, and Y.~Bengio.
\newblock Generative adversarial nets.
\newblock {\em arXiv:1406.2661}, 2014.

\bibitem{kim4}
H.~Kim and A.~Mnih.
\newblock Disentangling by factorising.
\newblock {\em arXiv:1802.05983v2}, 2018.

\bibitem{chen5}
R.~T.~Q. Chen, X.~Li, R.~Grosse, and D.~Duvenaud.
\newblock Isolating sources of disentanglement in {VAEs}.
\newblock {\em arXiv:1802.04942v1}, 2018.

\bibitem{gaussianvaes}
N.~Dilokthanakul, P.~A.~M. Mediano, M.~Garnelo, M.~C.~H. Lee, H.~Salimbeni,
  K.~Arulkumaran, and M.~Shanahan.
\newblock Deep unsupervised clustering with {G}aussian mixture variational
  autoencoders.
\newblock {\em arXiv:1611.02648v2}, 2017.

\bibitem{tomczak}
J.~M. Tomczak and M.~Welling.
\newblock {VAE} with a {VampPrior}.
\newblock {\em arXiv:1705:07120v5}, 2018.

\bibitem{chen4}
M.~Chen, X.~He, J.~Yang, and H.~Zhang.
\newblock 3-d convolutional recurrent neural networks with attention model for
  speech emotion recognition.
\newblock {\em IEEE Signal Processing Letters}, 25(10):1440--1444, 2018.

\bibitem{ridgeway}
K.~Ridgeway and M.~C. Mozer.
\newblock Learning deep disentangled embeddings with the f-statistic loss.
\newblock {\em Advances in Neural Information Processing Systems}, 2018.

\bibitem{doersch}
C.~Doersch.
\newblock Tutorial on variational autoencoders.
\newblock {\em arXiv:1606.05908v2}, 2016.

\bibitem{kingma}
D.~P. Kingma and M.~Welling.
\newblock Auto-encoding variational {B}ayes.
\newblock {\em arXiv:1312.6114v10}, 2014.

\bibitem{rezende2}
D.~J. Rezende, S.~Mohamed, and D.~Wierstra.
\newblock Stochastic backpropagation and approximate inference in deep
  generative models.
\newblock {\em arXiv:1401.4082}, 2014.

\bibitem{kulkarni}
T.~D. Kulkarni, W.~Whitney, P.~Kohli, and J.~B. Tenenbaum.
\newblock Deep convolutional inverse graphics network.
\newblock {\em arXiv:1503.03167v4}, 2015.

\bibitem{szabo1}
A.~Szabo, Q.~Hu, T.~Portenier, and P.~Favaro.
\newblock Challenges in disentangling independent factors of variation.
\newblock {\em arXiv:1711.02245v1}, 2017.

\bibitem{kingma3}
D.~P. Kingma, D.~J. Rezende, S.~Mohamed, and M.~Welling.
\newblock Semi-supervised learning with deep generative models.
\newblock {\em arXiv:1406.5298}, 2014.

\bibitem{dsprites}
L.~Matthey, I.~Higgins, D.~Hassabis, and A.~Lerchner.
\newblock {dSprites}: Disentanglement testing sprites dataset.
\newblock https://github.com/deepmind/dsprites-dataset/, 2017.

\bibitem{disentanglement}
C.~P. Burgess, I.~Higgins, A.~Pal, L.~Matthey, N.~Watters, G.~Desjardins, and
  A.~Lerchner.
\newblock Understanding disentangling in {Beta-VAE}.
\newblock {\em arXiv:1804.03599v1}, 2018.

\bibitem{hintondiagrams}
G.~E. Hinton and T.~Shallice.
\newblock Lesioning an attractor network: Investigations of acquired dyslexia.
\newblock {\em Psychological Review}, 98:74--95, 1991.

\bibitem{adam}
D.~P. Kingma and J.~L. Ba.
\newblock Adam: a method for stochastic optimization.
\newblock {\em arXiv:1412.6980v9}, 2017.

\bibitem{celebA}
Z.~Liu, P.~Luo, X.~Wang, and X.~Tang.
\newblock Deep learning face attributes in the wild.
\newblock {\em Proceedings of ICCV}, 2015.

\bibitem{openface2}
T.~Baltrusaitis, A.~Zadeh, Y.~C. Lim, and L-P. Morency.
\newblock {OpenFace} 2.0: Facial behavior analysis toolkit.
\newblock {\em 13th IEEE International Conference on Automatic Face and Gesture
  Recognition}, 2018.

\bibitem{theis}
L.~Theis, A.~{van den Oord}, and M.~Bethge.
\newblock A note on the evaluation of generative models.
\newblock {\em arXiv:1511.01844v3}, 2016.

\end{thebibliography}

\newpage

\section{Gated Variational AutoEncoders: Supplementary Material.}
This material supplements the results presented in the paper `Gated Variational AutoEncoder: Incorporating weak supervision to encourage disentanglement'. The results are provided for gated and un-gated versions of $\beta$-VAE, DIP-VAE-II and InfoVAE as well as for the Lasso and Random Forest (RF) regressors. The latent factors, $\mathbf{v}$, are $v_0=$ shape, $v_1=$  size, $v_2=$ rotation, $v_3=$ x-position and $v_4=$ y-position. Table \ref{table1} comprises the results for the informativeness of the inferred latent space, as measured using the Normalized Root Mean Squared Error (NRMSE), for each of the generative factors $\mathbf{v}$. It can be seen that gating improves the informativeness of the representation learned by the network, particularly when the downstream classification is performed by a random forest regressor. Table \ref{table2} comprises the results for the completeness of the inferred latent space, and shows how gating improves the completeness. Table \ref{table3} comprises the results for the disentanglement between \textbf{inferred} dimensions $\mathbf{z}$ of the latent space and demonstrates how disentanglement between the dimensions is improved with the use of gating. For all tables, bold font indicates an improvement of the gated version over the un-gated version. In summary, the incorporation of supervision with the gating of the gradients has improved completeness, informativeness, and disentanglement. Finally, Figure \ref{fig:simplediagram} shows the network architecture used for the dSprites dataset evaluation in detail. For these experiments, the latent space dimensionality $P$ is 8 (i.e. we have $z_0, z_1, ... z_7$).

\begin{table}[!h]
\centering
\caption{Inferred space \textbf{(un)informativeness} (lower is better) for each of the latent factors $\mathbf{v}$ averaged over 10 runs. Informativeness is the regressor NRMSE.}
 \label{table1}
 \resizebox{\linewidth}{!}{
\begin{tabular}{lllllllll}
\hline
\textbf{Regressor} & \textbf{Model} & \textbf{    } & $\mathbf{v}_0$ & $\mathbf{v}_1$  & $\mathbf{v}_2$ & $\mathbf{v}_3$ & $\mathbf{v}_4$ & \textbf{W. Avg.}  \\ \hline
%%%%%%%%%%%%%%%%%%%%%%%%%%%%%%%%%%%%%%%%%%%%%%%%%%%%%%%%%%%%%%%%%%%%%%%%%%%%%%%%%%%%%%%%%%
\multirow{6}{*}{Lasso} 
%%%%%%%%%%%%%%%%%%
% beta-VAE results
%%%%%%
&\multirow{2}{*}{$\beta$-VAE} & -- & 
0.933$\pm0.003$  & 0.597$\pm0.032$ & 0.997$\pm0.001$ &  0.457$\pm0.039$ &  0.465$\pm0.048$ &  0.690$\pm0.019$ \\
%%%%%%
& & \textbf{Gated} &
\textbf{0.395}$\pm0.19$ & \textbf{0.192}$\pm0.490$ & 1.000$\pm0.000$ & \textbf{0.295}$\pm0.175$ & \textbf{0.279}$\pm0.163$ & \textbf{0.432}$\pm0.087$ \\
%%%%%%%%%%%%%%%%%%
% InfoVAE results
%%%%%%
& \multirow{2}{*}{InfoVAE} & -- & 
0.956 $\pm0.003$  & 0.815$\pm0.026$ & 0.998$\pm0.000$ & 0.537$\pm0.020$ & 0.552$\pm0.028$ &  0.772$\pm0.013$ \\
%%%%%%
& & \textbf{Gated} & 
\textbf{0.363}$\pm0.083$  & \textbf{0.153}$\pm0.026$ & 1.000$\pm0.000$  & 0.410$\pm0.150$ & 0.479$\pm0.182$ & \textbf{0.481}$\pm0.062$ \\
%%%%%%%%%%%%%%%%%%
% Lasso:DIP-VAE-II results
%%%%%%
& \multirow{2}{*}{DIP-VAE-II} & -- &
0.927$\pm0.002$ & 0.443$\pm0.013$  &  0.999$\pm0.000$ & 0.441$\pm0.075$  & 0.453$\pm0.082$  &  0.653$\pm0.033$ \\
%%%%%%
& & \textbf{Gated} &
\textbf{0.780}$\pm0.098$ &  \textbf{0.368}$\pm0.072$ & 1.000$\pm0.000$ & 0.374$\pm0.255$ & 0.498$\pm0.252$ & 0.604$\pm0.086$ \\ \hline
%%%%%%%%%%%%%%%%%%%%%%%%%%%%%%%%%%%%%%%%%%%%%%%%%%%%%%%%%%%%%%%%%%%%%%%%%%%%%%%%%%%%%%%%%%
\multirow{6}{*}{\begin{tabular}[c]{@{}l@{}}Random\\ Forest\end{tabular}} 
%%%%%%%%%%%%%%%%%%
% beta-VAE results
%%%%%%
& \multirow{2}{*}{$\beta$-VAE} & -- & 
0.764$\pm0.023$  & 0.370$\pm0.010$ & 0.964$\pm0.004$ &  0.097$\pm0.005$ &  0.103$\pm0.012$ &  0.460$\pm0.007$ \\
%%%%%%
& & \textbf{Gated} &
\textbf{0.178}$\pm0.181$ & \textbf{0.065}$\pm0.043$ & 0.985$\pm0.004$ & \textbf{0.030}$\pm0.007$ & \textbf{0.032}$\pm0.009$ & \textbf{0.258}$\pm0.046$ \\
%%%%%%%%%%%%%%%%%%
% InfoVAE results
%%%%%%
& \multirow{2}{*}{InfoVAE} & -- & 
0.724$\pm0.027$  & 0.444$\pm0.014$ & 0.967$\pm0.003$ & 0.139$\pm0.011$ & 0.140$\pm0.011$ &  0.483$\pm0.004$\\
%%%%%%
& & \textbf{Gated} & 
\textbf{0.107}$\pm0.057$  & \textbf{0.044}$\pm0.010$ & 0.985$\pm0.004$  & \textbf{0.038}$\pm0.007$ & \textbf{0.038}$\pm0.007$ & \textbf{0.243}$\pm0.012$ \\
%%%%%%%%%%%%%%%%%%
% Lasso:DIP-VAE-II results
%%%%%%
& \multirow{2}{*}{DIP-VAE-II} & -- &
0.774$\pm0.012$ & 0.332$\pm0.008$  &  0.989$\pm0.008$ & 0.095$\pm0.018$  & 0.097$\pm0.017$  &  0.456$\pm0.008$ \\
%%%%%%
& & \textbf{Gated} &
\textbf{0.611}$\pm0.128$ &  \textbf{0.243}$\pm0.062$ & 0.991$\pm0.003$ & \textbf{0.059}$\pm0.018$ & 0.067$\pm0.033$ & \textbf{0.394}$\pm0.043$ \\ \hline
%%%%%%
\end{tabular}}
\end{table}

\begin{table*}[!h]
\centering
\caption{Inferred space \textbf{completeness} (higher is better) for each of the latent factors $\mathbf{v}$ averaged over 10 runs.}
 \label{table2}
  \resizebox{\linewidth}{!}{
\begin{tabular}{lllllllll}
\hline
\textbf{Regressor} & \textbf{Model} & \textbf{    } & $\mathbf{v}_0$ & $\mathbf{v}_1$  & $\mathbf{v}_2$ & $\mathbf{v}_3$ & $\mathbf{v}_4$ & \textbf{W. Avg.}  \\ \hline
%%%%%%%%%%%%%%%%%%%%%%%%%%%%%%%%%%%%%%%%%%%%%%%%%%%%%%%%%%%%%%%%%%%%%%%%%%%%%%%%%%%%%%%%%%
\multirow{6}{*}{Lasso} 
%%%%%%%%%%%%%%%%%%
% beta-VAE results
%%%%%%
& \multirow{2}{*}{$\beta$-VAE} & -- & 
0.240$\pm0.037$  & 0.239$\pm0.033$ & 0.238$\pm0.100$ &  0.330$\pm0.090$ &  0.332$\pm0.095$ &  0.276$\pm0.048$ \\
%%%%%%
& & \textbf{Gated} &
\textbf{0.463}$\pm0.118$ & \textbf{0.649}$\pm0.091$ & 0.219$\pm0.094$ & \textbf{0.532}$\pm0.243$ & \textbf{0.525}$\pm0.231$ & \textbf{0.478}$\pm0.104$ \\ 
%%%%%%%%%%%%%%%%%%
% InfoVAE results
%%%%%%
& \multirow{2}{*}{InfoVAE} & -- & 
0.152$\pm0.050$  & 0.140$\pm0.031$ & 0.106$\pm0.055$ & 0.211$\pm0.066$ & 0.172$\pm0.067$ &  0.156$\pm0.020$ \\
%%%%%%
& & \textbf{Gated} & 
\textbf{0.530}$\pm0.083$  & \textbf{0.662}$\pm0.026$ & \textbf{0.274}$\pm0.000$  & \textbf{0.514}$\pm0.150$ & \textbf{0.498}$\pm0.182$ & \textbf{0.495}$\pm0.062$ \\ 
%%%%%%%%%%%%%%%%%%
% Lasso:DIP-VAE-II results
%%%%%%
& \multirow{2}{*}{DIP-VAE-II} & -- &
0.374$\pm0.002$ & 0.444$\pm0.013$  &  0.165$\pm0.000$ & 0.388$\pm0.075$  & 0.361$\pm0.082$  &  0.346$\pm0.033$ \\
%%%%%%
& & \textbf{Gated} &
0.289$\pm0.117$ &  0.426$\pm0.170$ & 0.250$\pm0.117$ & 0.402$\pm0.228$ & 0.318$\pm0.195$ & 0.337$\pm0.081$ \\ \hline
%%%%%%
%%%%%%%%%%%%%%%%%%%%%%%%%%%%%%%%%%%%%%%%%%%%%%%%%%%%%%%%%%%%%%%%%%%%%%%%%%%%%%%%%%%%%%%%%%
\multirow{6}{*}{\begin{tabular}[c]{@{}l@{}}Random\\ Forest\end{tabular}} 
%%%%%%%%%%%%%%%%%%
%%%%%%%%%%%%%%%%%%
% beta-VAE results
%%%%%%
& \multirow{2}{*}{$\beta$-VAE} & -- & 
0.043$\pm0.022$  & 0.287$\pm0.061$ & 0.010$\pm0.006$ &  0.435$\pm0.092$ &  0.409$\pm0.090$ &  0.237$\pm0.032$ \\
%%%%%%
& & \textbf{Gated} &
\textbf{0.743}$\pm0.264$ & \textbf{0.947}$\pm0.058$ & 0.012$\pm0.004$ & \textbf{0.832}$\pm0.127$ & \textbf{0.805}$\pm0.131$ & \textbf{0.668}$\pm0.093$ \\
%%%%%%%%%%%%%%%%%%
% InfoVAE results
%%%%%%
& \multirow{2}{*}{InfoVAE} & -- & 
0.030$\pm0.012$  & 0.150$\pm0.029$ & 0.005$\pm0.0$ & 0.276$\pm0.110$ & 0.250$\pm0.096$ &  0.142$\pm0.026$ \\
%%%%%%
& & \textbf{Gated} & 
\textbf{0.721}$\pm0.055$  & \textbf{0.911}$\pm0.076$ & 0.011$\pm0.006$  & \textbf{0.772}$\pm0.102$ & \textbf{0.814}$\pm0.095$ & \textbf{0.646}$\pm0.032$ \\ 
%%%%%%%%%%%%%%%%%%
% Lasso:DIP-VAE-II results
%%%%%%
& \multirow{2}{*}{DIP-VAE-II} & -- &
0.245$\pm0.107$ & 0.569$\pm0.204$  &  0.054$\pm0.016$ & 0.449$\pm0.126$  & 0.413$\pm0.086$  &  0.346$\pm0.095$ \\
%%%%%%
& & \textbf{Gated} &
 0.285$\pm0.212$ &  \textbf{0.837}$\pm0.147$ & 0.004$\pm0.002$ & \textbf{0.721}$\pm0.161$ & \textbf{0.787}$\pm0.129$ & \textbf{0.527}$\pm0.077$ \\ \hline
%%%%%%
\end{tabular}}
\end{table*}

\begin{landscape}
 \begin{table*}[]
\centering
\small
\caption{\textbf{Disentanglement} (higher is better) for each of the inferred latent factors $\mathbf{z}$ averaged over 10 runs.}
 \label{table3}
\begin{tabular}{llllllllllll}
\hline
\textbf{Regressor} & \textbf{Model} & \textbf{    } & $\mathbf{z}_0$ & $\mathbf{z}_1$  & $\mathbf{z}_2$ & $\mathbf{z}_3$ & $\mathbf{z}_4$ & $\mathbf{z}_5$ & $\mathbf{z}_6$ & $\mathbf{z}_7$ & \textbf{W. Avg.}  \\ \hline
%%%%%%%%%%%%%%%%%%%%%%%%%%%%%%%%%%%%%%%%%%%%%%%%%%%%%%%%%%%%%%%%%%%%%%%%%%%%%%%%%%%%%%%%%%
\multirow{6}{*}{Lasso} 
%%%%%%%%%%%%%%%%%%
% beta-VAE results
%%%%%%
& \multirow{2}{*}{$\beta$-VAE} & -- & 
0.219$\pm 0.112$ & 0.250$\pm0.077 $ & 0.211$\pm0.107 $ & 0.230$\pm0.090 $ & 0.184$\pm 0.090$ & 0.264$\pm0.106 $ & 0.197$\pm0.107 $ &  0.262$\pm 0.137$ & 0.137$\pm0.039 $ \\
%%%%%%
& & \textbf{Gated} &
\textbf{0.556}$\pm0.158 $ & \textbf{0.595}$\pm0.179 $  & 0.237$\pm0.101 $ & 0.287$\pm0.125 $ & \textbf{0.472}$\pm 0.311$ & \textbf{0.566}$\pm0.289 $ & \textbf{0.652}$\pm0.278 $ &  \textbf{0.585}$\pm 0.244$ & \textbf{0.609}$\pm 0.136$  \\
%%%%%%%%%%%%%%%%%%
% InfoVAE results
%%%%%%
& \multirow{2}{*}{InfoVAE} & -- & 
0.261$\pm0.169 $  &  0.193$\pm 0.093$ & 0.206$\pm0.069 $ & 0.216$\pm0.081 $  & 0.196$\pm0.130 $ & 0.178$\pm0.102 $ &  0.310$\pm0.123 $ &  0.253$\pm0.091 $ & 0.240$\pm0.034 $ \\
%%%%%%
& & \textbf{Gated} & 
\textbf{0.578}$\pm0.091 $  & \textbf{0.605} $\pm0.116 $& 0.216$\pm0.099$ & 0.271$\pm0.099 $ & \textbf{0.542}$\pm0.259 $  &  \textbf{0.738}$\pm0.098 $ & \textbf{0.714}$\pm0.210 $ & \textbf{0.676}$\pm0.279 $  & \textbf{0.647}$\pm0.091 $ \\
%%%%%%%%%%%%%%%%%%
% Lasso:DIP-VAE-II results
%%%%%%
& \multirow{2}{*}{DIP-VAE-II} & -- &
0.299$\pm0.160$ & 0.248$\pm0.136 $ &  0.224$\pm0.123 $ & 0.278$\pm0.153 $  & 0.246$\pm0.169 $  &  0.284$\pm0.158 $ & 0.280$\pm0.125 $ &  0.220$\pm0.136 $ &  0.316$\pm0.113 $ \\
%%%%%%
& & \textbf{Gated} &
\textbf{0.479}$\pm0.150$   & \textbf{0.541}$\pm0.208 $  & 0.303$\pm0.120 $ & 0.354$\pm0.161 $ & 0.357$\pm0.180 $ & 0.439$\pm0.215 $  &  0.370$\pm0.208 $ & 0.451$\pm0.273 $  & \textbf{0.487}$\pm0.093 $
 \\ \hline
%%%%%%%%%%%%%%%%%%%%%%%%%%%%%%%%%%%%%%%%%%%%%%%%%%%%%%%%%%%%%%%%%%%%%%%%%%%%%%%%%%%%%%%%%%
\multirow{6}{*}{\begin{tabular}[c]{@{}l@{}}Random\\ Forest\end{tabular}} 
%%%%%%%%%%%%%%%%%%
% beta-VAE results
%%%%%%
& \multirow{2}{*}{$\beta$-VAE} & -- & 
0.226$\pm0.154 $ & 0.141$\pm0.099 $  & 0.248$\pm0.140 $ & 0.159$\pm0.129 $  & 0.301$\pm0.189 $ & 0.271$\pm0.139 $ &  0.296$\pm0.137 $ & 0.230$\pm0.133 $ &  0.172$\pm0.033 $ \\
%%%%%%
& & \textbf{Gated} &
\textbf{0.500}$\pm0.173 $ &  \textbf{0.511}$\pm0.150 $ & \textbf{0.901}$\pm0.133 $ & \textbf{0.880}$\pm0.132 $ & \textbf{0.810}$\pm0.131 $ &  \textbf{0.782}$\pm0.179 $ & \textbf{0.723}$\pm0.106 $ & \textbf{0.703}$\pm0.106 $ & \textbf{0.631}$\pm0.113 $ \\
%%%%%%%%%%%%%%%%%%
% InfoVAE results
%%%%%%
& \multirow{2}{*}{InfoVAE} & -- & 
0.129$\pm0.080 $  & 0.129$\pm0.068 $ & 0.129$\pm0.070 $ & 0.096$\pm0.056 $ & 0.078$\pm0.050 $ & 0.093$\pm0.053 $ & 0.133$\pm0.050 $ & 0.127$\pm0.042 $ & 0.113$\pm0.030 $ \\
%%%%%%
& & \textbf{Gated} & 
\textbf{0.490}$\pm0.103$  & \textbf{0.504}$\pm0.102 $ & \textbf{0.963}$\pm0.027 $ & \textbf{0.971}$\pm0.013 $ &  \textbf{0.766}$\pm0.137 $ & \textbf{0.737}$\pm0.126 $ & \textbf{0.650}$\pm0.091 $ &  \textbf{0.691}$\pm0.090 $ &  \textbf{0.632}$\pm0.047 $\\
%%%%%%%%%%%%%%%%%%
% Lasso:DIP-VAE-II results
%%%%%%
& \multirow{2}{*}{DIP-VAE-II} & -- &
0.221$\pm0.200 $ & 0.350$\pm0.202 $   &  0.297$\pm0.243 $ &  0.335$\pm0.241 $ & 0.378$\pm0.221 $ & 0.345$\pm0.222 $ & 0.344$\pm0.259 $ & 0.345$\pm0.238 $ & 0.197$\pm0.097 $ \\
%%%%%%
& & \textbf{Gated} &
\textbf{0.413}$\pm0.123 $  & 0.411$\pm0.141 $ & 0.540$\pm0.122 $  & 0.538$\pm0.149 $ &  \textbf{0.603}$\pm0.066 $ & 0.535$\pm0.115 $  & 0.553$\pm0.116 $ & 0.497$\pm0.152 $ &  \textbf{0.486}$\pm0.076 $ \\ \hline
\end{tabular}
\end{table*}

\begin{figure}
\centering
\includegraphics[width=0.85\linewidth,keepaspectratio]{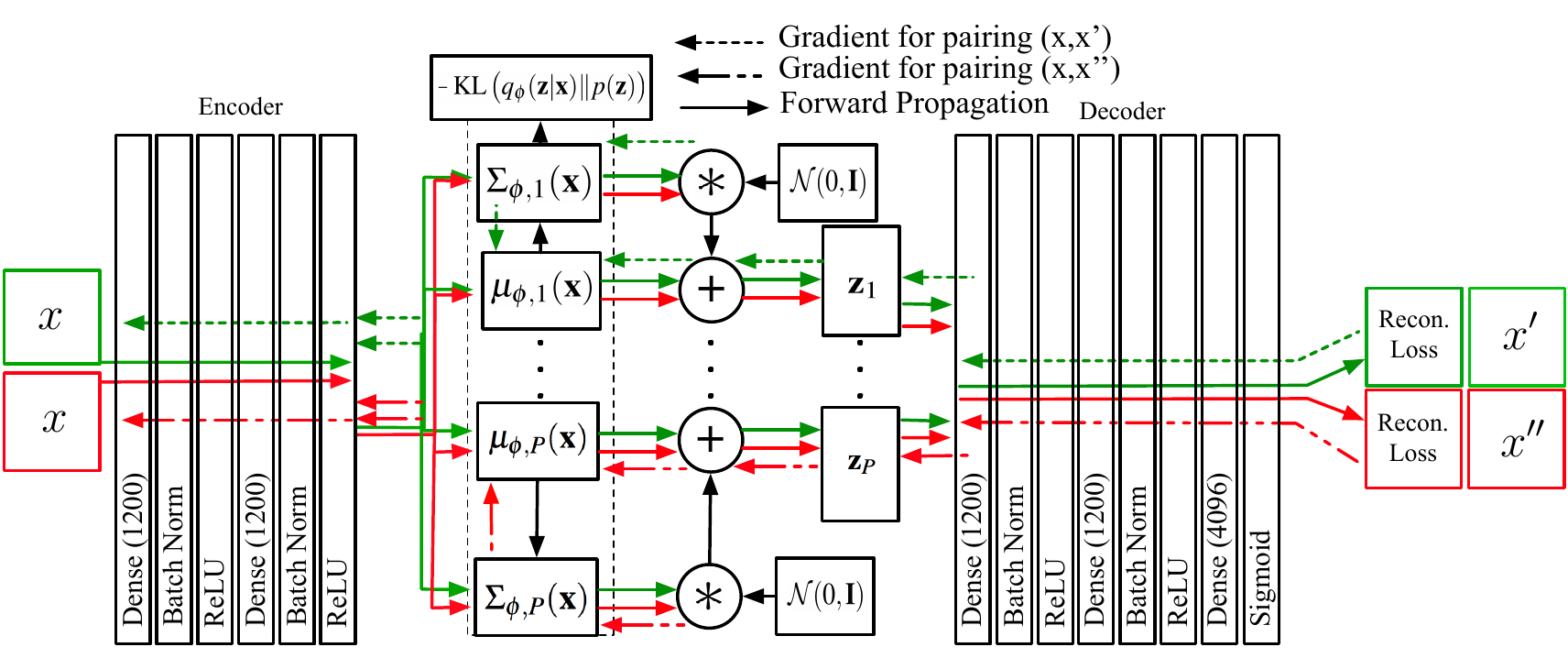}
\caption{Detailed overview of the Gated-VAE architecture. }
\label{fig:simplediagram}
\end{figure}
\end{landscape}

\end{document}